\documentclass[onesided,10pt]{article}                                                         % you want to use the \thanks command

\usepackage{subcaption}
\usepackage{color}
\usepackage{pdfpages}
\usepackage{cite}
\usepackage{algorithm}
\usepackage{algpseudocode}
\usepackage{amsmath}
\usepackage{multirow}
\usepackage{multicol}   %Columns
\usepackage[justification=centering]{caption}

\usepackage{graphicx}
\usepackage[bottom]{footmisc}

\usepackage{caption}
\usepackage{subcaption}

\usepackage{epstopdf,float,amsfonts}
\usepackage{epsfig}

\usepackage{todonotes} % for comments
 \title{\LARGE \bf Enhancing Cyber Resilience of Networked Microgrids using Vertical Federated Reinforcement Learning}
\author{\small{Sayak Mukherjee$^{1*}$, Ramij R. Hossain$^{1,2*}$, Yuan Liu$^{1*}$, Wei Du$^{1}$, Veronica Adetola$^{1}$,} \\ \small{Sheik M. Mohiuddin$^{1}$, Qiuhua Huang$^{1}$, Tianzhixi Yin$^{1}$, Ankit Singhal$^{1}$}
\\ \small{Pacific Northwest National Laboratory$^{1}$, USA, Iowa State University$^{2}$, USA}
\\ \small{Corresponding email: sayak.mukherjee@pnnl.gov, 
*Equal contributions.}

\thanks{The research described in this paper is part of the Resilience through Data-Driven, Intelligently Designed Control Initiative (RD2C) at Pacific Northwest National Laboratory (PNNL). It was conducted under the
Laboratory Directed Research and Development Program at PNNL, a multiprogram national laboratory operated by Battelle for the U.S. Department of Energy. Q. Huang was with PNNL. R. R. Hossain was a student intern at PNNL. 
}
\vspace{-0.3in}
}
\begin{document}
\normalsize
\pagestyle{empty}
\date{}
\maketitle
\thispagestyle{empty}
\begin{abstract}
This paper presents a novel federated reinforcement learning (Fed-RL) methodology to enhance the cyber resiliency of networked microgrids. We formulate a resilient reinforcement learning (RL) training setup which (a) generates episodic trajectories injecting adversarial actions at primary control reference signals of the grid forming (GFM) inverters and (b) trains the RL agents (or controllers) to alleviate the impact of the injected adversaries. To circumvent data-sharing issues and concerns for proprietary privacy in multi-party-owned networked grids, we bring in the aspects of federated machine learning and propose a novel Fed-RL algorithm to train the RL agents. To this end, the conventional horizontal Fed-RL approaches using decoupled independent environments fail to capture the coupled dynamics in a networked microgrid, which leads us to propose a multi-agent vertically federated variation of actor-critic algorithms, namely federated soft actor-critic (FedSAC) algorithm. We created a customized simulation setup encapsulating microgrid dynamics in the GridLAB-D/HELICS co-simulation platform compatible with the OpenAI Gym interface for training RL agents. Finally, the proposed methodology is validated with numerical examples of modified IEEE 123-bus benchmark test systems consisting of three coupled microgrids.      
\end{abstract}

\newpage
\section{Introduction}
\label{sec:intro}
 {\color{black} To achieve net-zero-energy by 2050 \cite{netzeroDOE}}, networked microgrids are one of the sought-after solutions for self-sustaining power grids that can deliver and manage efficient renewable integration. These resources are interfaced with the power grids using power-electronic devices, specifically converter/inverter technologies. Current research works in inverter design have led to the prominent rise of grid-forming inverters (GFMs) \cite{du2020modeling}. GFMs can act as a controllable voltage source behind a coupling impedance and can directly control the voltage and frequency of the microgrid. The control layers in the GFM-based microgrid setting consists of multiple layers, ranging from primary to higher-level control layers \cite{bidram2012hierarchical, guerrero2010hierarchical,singhal2022consensus}. These controllable resources are therefore subjected to various resiliency concerns where the power electronic interfaces may get subjected to adversarial cyber attacks. Literature on various possibilities of cyber events for microgrids, and resiliency aspects can be found in references \cite{ deng2020distributed,zhou2020cyber, sahoo2020resilient}. In multi-party ownership models, different zones in a networked microgrid can be owned by different utilities/operators with limited data and proprietary information exchange during operation. Moreover, with the increasing complexity of microgrid operations and due to various modeling uncertainties, exact knowledge of the dynamics is difficult to acquire. Therefore, we are intrigued to ask the following two major questions -- \textit{how can we design higher-level controllers with limited knowledge about the networked microgrids that can inject resiliency into the microgrid operations? And how to tackle the limited data sharing issue across the networks of microgrids considering the dynamic electrical coupling?}    

Reinforcement learning has seen considerable progress over the last few years, where complex dynamic tasks are solved in a Markov decision process (MDP)-based framework using interactions with the environment using value-based or policy gradient-based or a combination of these approaches in works such as  \cite{LillicrapHPHETS15, SAC}, to name a few. These learning control problems face challenging bottlenecks when the objective is to optimize over multiple agents in a coupled dynamic environment with segregated action and state spaces, leading to the research of multi-agent RL such as \cite{marl1_review, marl4_maddpg}. RL has been used for voltage control \cite{others_voltage}, control of energy storage in microgrids \cite{others_micro}, wide-area damping control \cite{others_wadc}, volt-VAr control in distribution grids \cite{others_voltvar}, etc. Moreover, the learning problem becomes more involved if we impose privacy constraints. One promising solution is federated learning (FL) \cite{bonawitz2019towards, li2020federated}, which shares model parameters and gradients between the zones or entities instead of sharing new input data. To this end, Fed-RL \cite{qi2021federated, wang2020federated, zhuo2019federated} is at the nascent stage, and there are few recent Fed-RL applications in power systems such as for decentralized volt-var control \cite{liu2022federated}, and energy management system for smart homes \cite{lee2020federated}. 
% Among power system applications, \cite{liu2022federated} presented a decentralized volt-var control problem using Fed-RL based on power-flow based study (without considering any microgrid dynamics). A federated reinforcement learning-based solution for energy management system for smart homes is studied in \cite{lee2020federated}. 
In contrast, our paper considers a holistic design architecture for multi-party owned networked (coupled) microgrids and presents \textit{vertically} federated reinforcement learning approach to achieve a resilient control architecture in mitigating the impacts of adversarial actions at the reference signals of primary control loops of the grid-forming inverters (GFM). We tested our proposed methodology in a practical benchmark module, the IEEE-123 bus test feeder, with three microgrids having individual GFM inverters.
\\
\noindent \textbf{Contributions.}
We summarize the contributions of this paper as follows:
\\
1. We tackle the problem of resilient control of networked microgrids by a purely learning-driven multi-agent adversarial reinforcement learning approach, thereby mitigating the requirement of accurate dynamic model knowledge in control design. \\
2. We bring the ideas of federated learning into the coupled multi-agent reinforcement learning designs to address the data-sharing concerns among different microgrid owners, and this leads to a novel Fed-RL algorithm, FedSAC.\\
3. We created a novel software module, \textit{Resilient RL Co-simulation Platform for Microgrids}, compatible with the OpenAI Gym \cite{brockman2016openai} interface for the application of any benchmark RL methods. Grid simulator GridLAB-D \cite{chassin2014gridlab} and HELICS \cite{palmintier2017design} co-simulation platforms are utilized in the developed module to perform necessary dynamic simulations of microgrids.

% \begin{enumerate}
%     \item We showcase the problem of resilient control of networked microgrids using the grid-forming inverters as our actuation agent using a purely learning-driven approach using multi-agent reinforcement learning, thereby mitigating the requirement of accurate dynamic model knowledge. 
%     \item We bring forth the ideas of federated learning into the coupled multi-agent reinforcement learning designs in response to the data sharing concerns by different microgrid ownership leading to a novel vertical federated reinforcement learning design {\color{black} utilizing state-of-the-art soft actor-critic (SAC) algorithm.}
%     \item Our numerical experiments were performed in a practical benchmark module, namely, the IEEE-123 bus test feeder with three microgrids where each of the microgrids contains one grid-forming inverter. We have created a novel software setup where the microgrid co-simulation platform using GridLAB-D and HELICS are integrated into an OpenAI Gym interface for subsequent RL experiments.
% \end{enumerate}

\section{Resilient Reinforcement Learning Problem}
% \section{Preliminaries: Microgid dynamics, Resilient RL formulation}
\label{sec.ARS}

\noindent {\bf Microgrid dynamics:} We consider $r$ number of coupled microgrids, and a total of $N$ number of GFM inverters and $M$ number of buses in the network. For the $i^{th}$ GFM inverter, it can be modeled as an AC voltage source with internal voltage $E_i$, and phase angle $\delta_i$ mathematically represented as,
    $\dot{\delta}_i = u_i^\delta,
    E_i = u_i^V.$
Here $u_i^\delta, u_i^V$ are the frequency and voltage control input signals or reference signals to the inverter. The primary control of the GFM inverters constitutes the droop controls as follows,
\begin{align}
    \omega_i^{ref} = \omega_i^{nom} - m_{Pi} (P_i - P_i^{set}),\\
    V_i^{ref} = V_i^{set} - m_{Qi} (Q_i - Q_i^{nom}),
\end{align}
where frequency control input $u_i^{\delta}$ is equal to the reference frequency $\omega_i^{ref}$, i.e. $u_i^{\delta}=\omega_i^{ref}$ and voltage control input $u_i^V$ is obtained by passing $V_i^{ref} - V_i$ through a proportional-integral (PI) regulator. Here voltage, active power, and reactive power are denoted as $V_i, P_i$, and $Q_i$ respectively, with droop gains as $m_{Pi}$, and $m_{Qi}$. The purpose of primary control is to maintain proportional power sharing and prevent circulating reactive power. However, due to its proportional nature, accurate regulation can not be achieved as they always end up with a steady state deviation in both frequency and voltage. Many times, secondary control is needed to accomplish the removal of these errors. For more details please see \cite{du2020modeling} 

\noindent {\bf Resilient RL Aspects:} The resilient RL controls will act like a supervisory control layer on top of existing primary and secondary controls if present. The target here is to train the RL agent in presence of cyber vulnerabilities. Let the RL outputs be denoted as $P_i^{res}, V_i^{res}$, i.e., the resilient control inputs. Thereafter these higher-level control signals will be added to the nominal/pre-specified set-points $P_{i-nom}^{set}$, and $V_{i-nom}^{set}$ such that for the $i^{th}$ GFM inverter we have,
\begin{align}
    P_i^{set} = P_{i-nom}^{set}+P_i^{res},
    V_i^{set} = V_{i-nom}^{set}+ V_i^{res}.\label{eq5}
\end{align}
The concatenated RL control inputs $u^{res} = [P_i^{res}, V_i^{res}]_{i=1,..,N}$ will be designed as the feedback function of the microgrid observations ($O$), discussed later, passed through a function $f(\cdot)$ with parameters $\theta$ such that $u^{res} = f_\theta(O)$. During training, inverter attack is emulated by adding attack signals at randomly selected $i^{th}$ inverter perturbing the active power and voltage set-points. 
% For a particular episode of the training, if the $i^{th}$ inverter is attacked, we will add attack signals to the active power and voltage set-points such that, 
\begin{align}
    P_i^{set} = P_{i-nom}^{set}+P_i^{res} + P_i^{attack},\\
    V_i^{set} = V_{i-nom}^{set}+ V_i^{res} + V_i^{attack},
\end{align}
% and similarly other episodes will cover the remaining set of attacked and healthy inverters. 
Therefore, we need to design resilient controllers to mitigate the effects of such adversaries. Rule-based design of such controllers is not feasible due to stochastic nature of the problem. This motivated to the RL-based controller design which is achieved by casting the resilient microgrid control problem in a (partially observable) MDP setting, 
% Next, we re-frame the above dynamics into the  RL formulation for the resilient microgrid control with a (partially observable) MDP 
defined by a tuple $(S,A, \mathcal{P},r,\gamma)$ \cite{RL}. The state space (microgrid dynamics) $S \subseteq \mathbb{R}^n$ and action space (through GFM inverter setpoints) $ A \subseteq \mathbb{R}^m$ are continuous, environment transition function $\mathcal{P}: S \times A \to S$  characterizes the stochastic transition of the microgrid states during the dynamic events along with reward $r: S \times A \to R$, and the discount factor $\gamma \in (0,1)$. \textit{Observation space:} Although complex microgrid dynamics consist of many differential and algebraic variables, we focus on a partial set of such variables depending on the underlying problem. Without loss of generality, considering attacks at voltage set-points, 
% it is physically not possible to measure all the states, and one can only measure a partial set of states.
bus voltage magnitudes $V_i(t)$ (Note that $V_i(t)$ are different from inverter voltage set points in (\ref{eq5})) are
% measurable, and we construct our
taken as the observation variables $O$.
% using these. 
\textit{Action space:} The RL agents can implement their actions using the $P_i^{res}$, and $V_i^{res}$ inputs for each individual grid forming inverters in a continuous manner, however, practical set-point limiters are implemented to keep the inputs within tolerable bounds.
% {\it Recovery bounds:} 
{\it Rewards:} The resilient design needs to keep the quality of service (QoS) variables within desired bounds. 
% For this design, the operator would like to keep the terminal bus voltages of the inverter close to nominal steady state voltage. 
For this, the reward $r(t)$ at time $t$ is defined as follows:
% within $1\%$ deviation from the nominal steady state values $V_{{i,ss}}$ of respective bus $i$. Please note we make it more conservative than the practical requirement of $5\%$ margin. 

%************** Don't remove this space********************************** \small is used

\footnotesize
\vspace{-0.1in}
\begin{gather}
r(t) =
\begin{cases}
 - cu_{ivld} \;\; \text{if}\;\; t \leq t_a,\\
   - \sum_i Q_i {\lvert \lvert V_i(t) - V_{{i,ss}}\rvert \rvert}_2, \;\; \text{if}\;\; t > t_a \;\;\text{and}\;\;\Big\{V_i(t) < 0.99 V_{{i,ss}} \nonumber \\ \;\;\;\;\;\;\;\;\;\;\;\; \text{or} \;\;V_i(t) > 1.01 V_{{i,ss}}\Big\},\\
     0, \;\; \text{if}\;\; t > t_a \;\;\text{and}\;\; \Big\{0.99 V_{{i,ss}}\leq V_i(t) \leq 1.01 V_{{i,ss}}\Big\}.
\end{cases}
\end{gather}
\normalsize
where, $t_a$ is the instant of the adversarial action, $V_i(t)$ is the voltage magnitude for bus $i$ in the power grid at time $t$, and $V_{i,ss}$ is the steady-state voltage of bus $i$ before the attack, $u_{ivld}$ is the invalid action penalty if the DRL agent provides action when the network is not attacked. $Q_i$ and $c$ are weights corresponding to voltage deviation and invalid action penalty, respectively.

% \setlength{\textfloatsep}{0.1cm}
% \setlength{\floatsep}{0.1cm}
% \footnotesize
% \begin{algorithm}[t]
% \caption{Adversarial training overview for Resilient RL of networked microgrids}
%     \begin{algorithmic}[1]
%     \State Initialize policies for different RL controllers actuated using GFM inverters.

\begin{figure}[t]
  \centering
    \includegraphics[width=\textwidth]{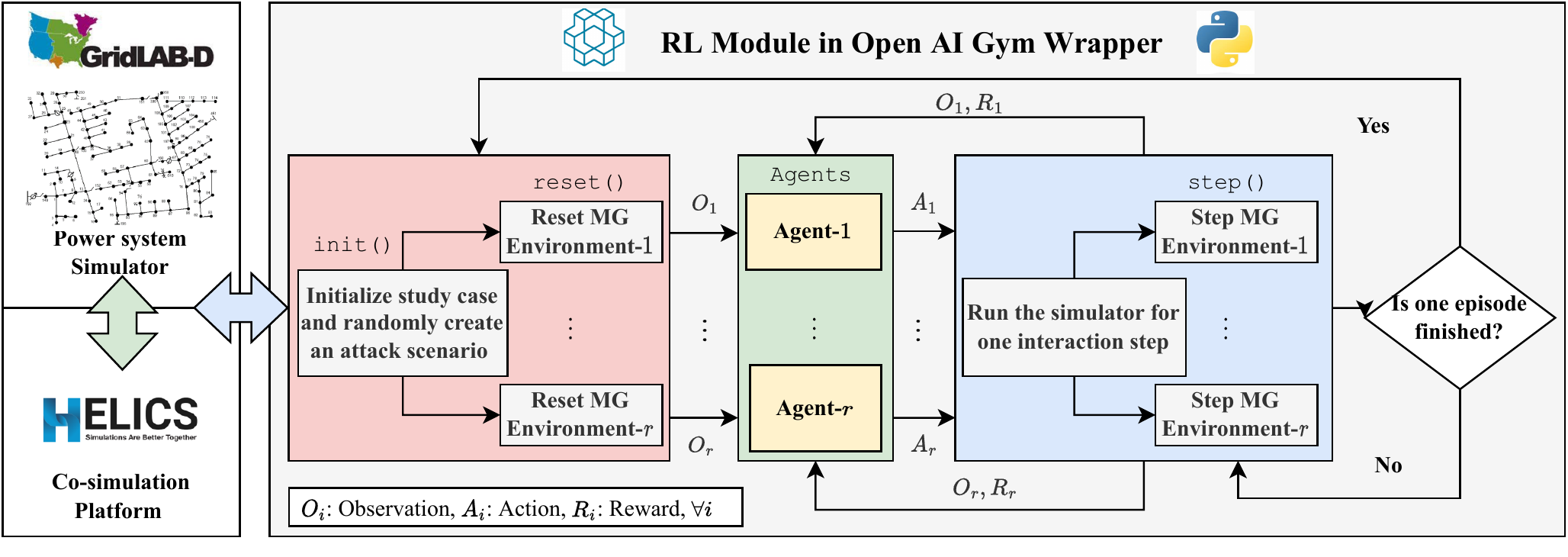}
  \caption{\small{Resilient RL Co-simulation Platform for Microgrids}}
  \label{f11}
 \end{figure}

% \State Create a pool of adversarial scenarios mimicing attacks at the reference set-points of active power and voltages.
% 	\For {$eps = 1,2,\dots,n_f$} 
% 	    \State \textbf{Sample} an adversarial run-time scenario.
%         \State \textbf{Generate} episodic trajectory data with GridLAB-\newline\hspace*{1.4em}D/HELICS co-simulation with OpenAI Gym.
%         % \newline\hspace*{1.4em} OpenAI Gym.
%         \State \textbf{Pass} global reward functions to each controllers.
%         \State \textbf{Communicate} between different controllers (multi-\newline\hspace*{1.4em}agent/federated approach) to update individual policy\newline\hspace*{1.4em}parameters.
% % \newline\hspace*{1.4em} 
%         \EndFor
% % \normalsize
%     \end{algorithmic} 
% \label{alg_overview}
% \end{algorithm}
% \normalsize
% \setlength{\textfloatsep}{0.1cm}
% \setlength{\floatsep}{0.1cm}
\noindent \textbf{Resilient RL Co-simulation Platform for Microgrids:}
% Current research trends in the RL community use the benchmarking algorithms created using the OpenAI Gym platform. The development of such a platform requires a simulation engine to be wrapped under the Python API interface for the use of operators. Our simulation architecture uses GridLAB-D as the network microgrid simulation engine; however, as we are interested in the control tasks, we need to use GridLAB-D's subscription/publication architecture to manipulate a few control set-points using externally written Python codes. This is enabled by the co-simulation platform HELICS \cite{palmintier2017design}.
We developed an OpenAI Gym compatible co-simulation platform for microgrids which is suitable to train any standard benchmark RL algorithms. The microgrid dynamics are simulated with the distribution grid simulator GridLAB-D. But, developing such a platform requires a simulation engine wrapped under the Python API interface. As we are interested in the control tasks, we need to use GridLAB-D's subscription/publication architecture to manipulate a few control set-points using externally written python codes. This python wrapper design for GridLab-D is achieved using the HELICS co-simulation platform. Therefore, our architecture uses two main modules, (a) the Microgrid Co-simulation module enabled by the GridLAB-D/HELICS engine and (b) the Resilient RL algorithmic development (described in the next section) module compatible with the OpenAI Gym. The OpenAI Gym environment uses few standard functions such as \texttt{init(), reset(), step()}, where the function \texttt{init()} initializes the power flow cases, attack instant, attack duration, observation, action, and other necessary variables. The function \texttt{reset()} randomly selects necessary configurations including adversarial attacks and interacts with the GridLAB-D/HELICS module to start a trajectory roll-out. Next, the \texttt{step()} function is called to establish the agent interaction with the GridLAB-D/HELICS dynamics. At each step, (a) agent actions are passed to the system, and (b) resulting observations and rewards are returned to the RL module. We created a pool of adversarial scenarios mimicking actuation attacks at the inner-control (primary control) loop of GFMs. In each episodic run, these adversarial events are selected randomly, and the resultant episodic trajectory information, including observations, actions, and rewards, is sent to the RL module for training of the resilient RL agent. The detailed framework is shown in Fig. \ref{f11}, which we utilized for training of our proposed Fed-RL agents.

\section{Resilient Vertical Fed-RL}
\setlength{\textfloatsep}{0.1cm}
\setlength{\floatsep}{0.1cm}
% \footnotesize
\begin{algorithm}[t]
\caption{Vertical Federated Resilient RL of networked microgrids}
\begin{algorithmic}[1]
\State \textbf{Initialize} critics and policies for different RL controllers actuated using GFM inverters, i.e., $Q^\alpha_{\phi_\alpha}$, and $\pi^\alpha_{\theta_\alpha}$ for the microgrid $\alpha$.

\For {$eps = 1,2,\dots,n_f$} 
	    \State \textbf{Sample} an adversarial run-time scenario from the \newline \hspace*{1.4em} adversarial action pool.

    \State \textbf{Generate} episodic trajectory data with the GridLAB-\newline \hspace*{1.4em} D/HELICS-OpenAI Gym emulator.
    
    \State For each of the MG $\alpha$, \textbf{use} $o_\alpha$, and $u^{res}_\alpha$ from the \newline \hspace*{1.4em}MG $\alpha$ to update the critic $Q$-function networks $Q^\alpha_{\phi_\alpha}$,\newline \hspace*{1.4em} $\alpha = 1,..,r$. 
    % \newline \hspace*{1.4em} 
    \State \textbf{Send} critic $Q$-function network $Q^\alpha_{\phi_\alpha}$ models to the \newline \hspace*{1.4em}central  coordinator or the grid operator.
    
    \State \textbf{Perform} information fusion at the coordinator by \newline \hspace*{1.4em}an averaging operation, and return the aggregated\newline \hspace*{1.4em} critic network model $Q_\phi$ to each microgrid.
    % \newline \hspace*{1.4em}  

    \State \textbf{Perform} gradient updates on the policy parameters \newline \hspace*{1.4em}of $\pi^\alpha_{\phi_\alpha}$ for each MG using the local 
    % \newline \hspace*{1.4em} 
    observations, \newline \hspace*{1.4em}actions and the global critic network model $Q_\phi$. 
\EndFor
\normalsize
\end{algorithmic} 
\label{alg_fedRL}
\end{algorithm}
\setlength{\textfloatsep}{0.1cm}
\setlength{\floatsep}{0.1cm}
For $r$ number of coupled microgrids, the $\alpha^{th}$ microgrid's actions (implemented using GFM inverters) and the observations consisting of the terminal bus voltages in a concatenated way are denoted as $u_\alpha^{res}$, and $o_\alpha$, respectively. Unlike conventional federated RL, also referred to as horizontal Fed-RL \cite{qi2021federated}, the $\alpha^{th}$ microgrid environment is not independent of the $\beta^{th}$ microgrid environment because of the electrical network coupling. Moreover, we have $\cup_{\alpha} u_{\alpha}^{res} = u^{res}, \cup_{\alpha} o_{\alpha} = O $ for the global networked microgrid environment. We will have $r$ number of policies of the form $u_{\alpha}^{res} = \pi^\alpha (o_\alpha)$, $\alpha = 1,..,r$. We consider deep neural network parametrized policies, therefore, denoting by $\pi^{\alpha}_{\theta_\alpha}(.)$ with the parameters $\theta_\alpha$ for microgrid $\alpha$. We propose using an actor-critic reinforcement learning architecture to infuse the federated learning characteristics where the critic Q-networks for microgrid $\alpha$ denoted as $Q^\alpha_{\phi_\alpha}$ with the neural network parameters $\phi_\alpha$. \par
\begin{figure}[t]
  \centering
    \includegraphics[width=\textwidth]{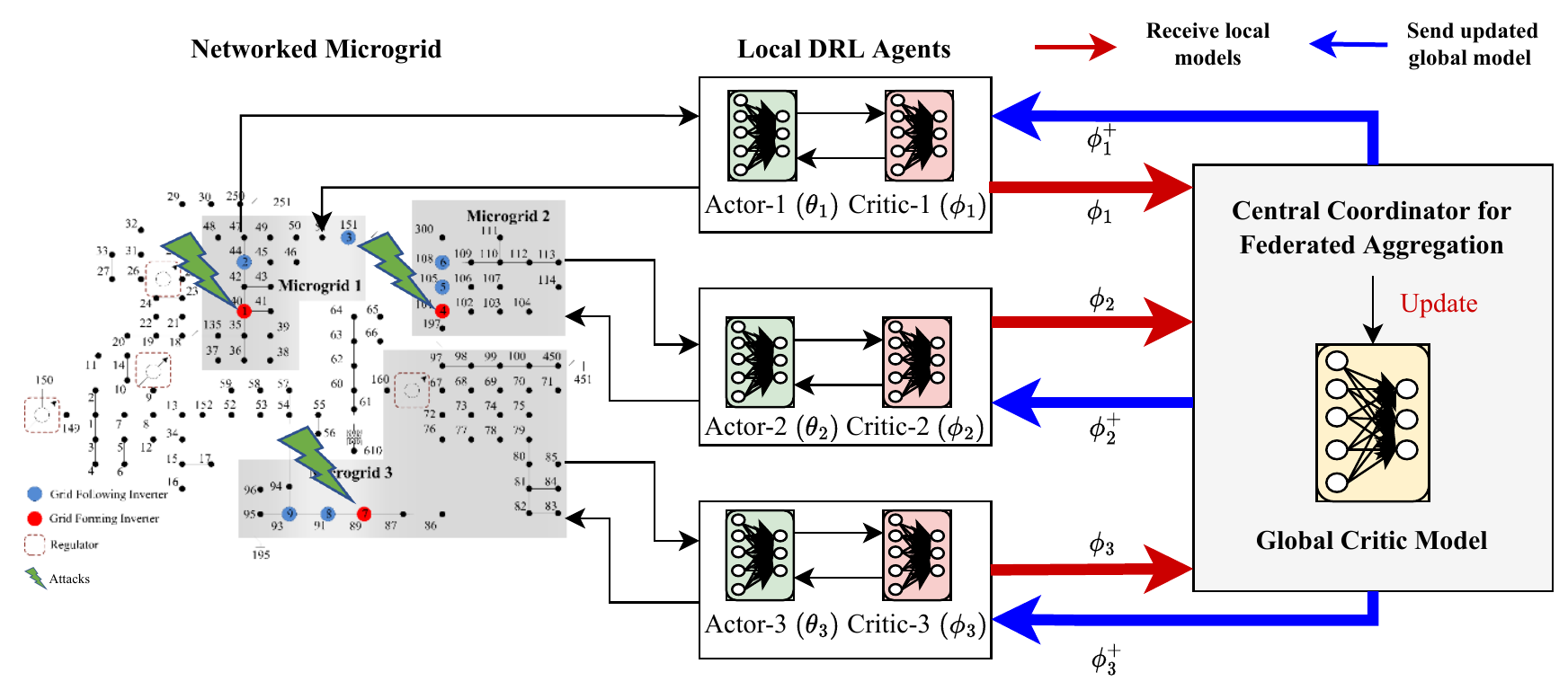}
  \caption{\small{Fed-RL framework for Networked Microgrid}}
  \label{f12}
    %\vspace{-0.1 in}
 \end{figure}

   
The training of DRL algorithms for real-world complex dynamics (e.g., microgrid problem) is nontrivial. Moreover, federated learning poses significant challenges in achieving efficient and stable training of the RL agents. The federated aspects are integrated into a multi-agent RL setting. We use microgrid-wise decentralized observation and action spaces and initially update the critic networks $Q^\alpha_{\phi_\alpha}$ with local data. After that, we propose to send these critics models to the coordinator, which can be implemented at the operator control center of the networked microgrids. Sufficient encryption/privacy-preserving techniques can be utilized to prevent model parameter leakage. Subsequently, these critics models are aggregated to infuse the influence from different microgrid's dynamic behaviors, and an aggregated model is created which is then transferred back to individual microgrid agents. This aggregated critic is then used to update the microgrid policies again using the local data. This strategy presents a novel multi-agent decentralized implementation architecture where the influence of the dynamics of the other coupled environments is captured by the federated averaging of the critic networks. Moreover, the intricacies of the federated aspect are maintained by leveraging local data usage and only sharing critic models with the coordinator. Alg. \ref{alg_fedRL} presents the main steps, and an overview of the comprehensive framework is given in Fig. \ref{f12}. However, during implementation, we extend these ideas to the state-of-the-art SAC algorithm \cite{SAC} with entropy regularization, and the policy is trained to maximize a trade-off between expected return and entropy, a measure of randomness in the policy. We modified standard SAC algorithm from Stable Baselines \cite{raffin2021stable} to incorporate federated learning framework. The resulting algorithm is presented in Algorithm \ref{alg_fedSAC}.
% \vspace{-0.05in}

The observation space $o_\alpha$ of individual microgrid agents contains the voltages of different buses, and due to network constraints, the steady-state voltage varies over buses. Therefore, the $o_\alpha$ is not necessarily identical for different agents $\alpha$. Consequently, the state-action space for individual critics and target critics differ or, in other words, follow different distributions. Hence, averaging the critic and target critic at step 18 in Alg. \ref{alg_fedSAC} get affected due to this distribution shift. To circumvent, we normalize the observation space for individual microgrids with respect to their steady-state values. On the other hand, the standard SAC algorithm concurrently learns a policy $\pi_{\theta_\alpha}^\alpha$ and two Q-functions $Q^\alpha_{\phi_\alpha^1}, Q^\alpha_{\phi_\alpha^2}$ following Clipped Double Q-trick. Clipped Double Q-learning is a variant on Double Q-learning that upper-bounds the less biased Q estimate $Q^\alpha_{\phi_\alpha^1}$ by the biased estimate $Q^\alpha_{\phi_\alpha^2}$, and this is achieved by taking a minimum over two Q estimates in step-12 and step-14 of Alg. \ref{alg_fedSAC}. In general, this Clipped Double Q-trick helps improve learning in standard SAC. But, in our initial experiments, we observed that even after promising performance at the initial stage of the training, the FedSAC algorithm can face stabilization issue in the training performance due to federated averaging. The weight averaging of the critic and target network (at step 18 of Alg. \ref{alg_fedSAC}) and subsequent minimization operation (step-12 and step-14 of Alg. \ref{alg_fedSAC}) can be detrimental for actor update. Again, it is also observed that for the initial part of training (when actor and critic networks are not trained), there is a need for utilizing the Clipped Double Q-trick. To overcome this issue, for the first half of iterations we followed the Clipped Double Q-learning, after that we select only one critic/target pair either $\{{\phi_{\alpha}^{1}},{\phi_{\alpha}^{\text{tar},1}}\}$ or $\{{\phi_{\alpha}^{2}},{\phi_{\alpha}^{\text{tar},2}}\}$ for federated averaging and actor-critic update. 

\footnotesize
\begin{algorithm}[H]
\caption{Federated Soft Actor Critic (FedSAC)}
\begin{algorithmic}[1]
\State Initialize environments $e_\alpha$, policy $\pi^\alpha_{\theta_\alpha}$ with parameters $\theta_{\alpha}$, critic $Q^\alpha$ with parameters $\phi^1_\alpha,\phi^2_\alpha$, and empty replay buffer $\mathcal{D}_{\alpha}$ for all $\alpha = 1,\cdots,r$
\State Set target critic parameters ${\phi_{\alpha}^{\text{tar},1}} \leftarrow \phi^1_\alpha, {\phi_{\alpha}^{\text{tar},2}} \leftarrow \phi^2_\alpha$ for all $\alpha = 1,\cdots,r$.
\Repeat  
\State Observe $o_{\alpha}$, and select action $u_{\alpha}^{res} \sim \pi^{\alpha}_{\theta_{\alpha}}(\cdot|o_\alpha)$ for \newline \hspace*{1.4em}all $\alpha = 1,\cdots,r$.
% \EndFor
\State Concatenate actions and form $\cup_{\alpha} u_{\alpha}^{res} = u^{res}$. 
\State Execute and observe next state $o'_{\alpha}$, reward $r_{\alpha}$, and \newline \hspace*{1.4em}done signal $d_{\alpha}$ for all $\alpha = 1,\cdots,r$.
\State Store $(o_\alpha,u_{\alpha}^{res},r_\alpha,{o'}_\alpha,d_\alpha)$ in replay buffer $\mathcal{D}_\alpha$ for \newline \hspace*{1.4em}all $\alpha = 1,\cdots,r$.
\State If ${\cap_{\alpha} d_\alpha} \rightarrow$ TRUE, reset environment state.
\If {Update step is True}
\For {$\alpha = 1,2,\dots,r$} 
 \State Randomly sample a batch of transitions,  \newline \hspace*{4.4em}$B_{\alpha} = \{(o_\alpha,u_{\alpha}^{res},r_\alpha,{o'}_\alpha,d_\alpha)\}$ from $\mathcal{D}_{\alpha}$.
 \State Compute targets for the $Q^\alpha$ functions, (where\newline \hspace*{4.4em} $\tilde{u}_{\alpha}^{res} \sim \pi^\alpha(\cdot|o'_\alpha)$)
   \vspace{-0.05in}
   
  \small
 \begin{equation} \nonumber 
     y_\alpha = r_\alpha + \gamma(1-d_\alpha) \Big ( \min_{i=1,2} Q^\alpha_{\phi_{\alpha}^{\text{tar},i}}(o'_\alpha,\tilde{u}_{\alpha}^{res}) - \zeta \log \pi^\alpha(\tilde{u}_{\alpha}^{res}|o'_\alpha) \Big )
    %  \;\;, \;\;\tilde{u}_{\alpha}^{res} \sim \pi^\alpha(\cdot|o'_\alpha).
 \end{equation}
 \normalsize
 \State Update $Q^\alpha$ functions using: 
   \vspace{-0.05in}
  \small
 \begin{equation}\nonumber 
 \nabla_{\phi_\alpha^i} \frac{1}{|B|} \sum_{(o_\alpha,u_{\alpha}^{res},r_\alpha,{o'}_\alpha,d_\alpha)\in B_\alpha} \Big (Q^\alpha_{\phi_\alpha^i}(o_\alpha,u_{\alpha}^{res})-y_\alpha\Big)^2,i = 1,2.
 \end{equation}
  \normalsize
  \State Update policy:
  \vspace{-0.05in}
  \small
  \begin{multline} \nonumber 
  \nabla_{\phi_\alpha^i} \frac{1}{|B|} \sum_{o_\alpha\in B_\alpha} \min_{i=1,2} \Big(Q^\alpha_{\phi_\alpha^i}(o_\alpha,\tilde{u}^{res}_{\theta_\alpha}(o_\alpha)) - \\ \zeta \log \pi^\alpha(\tilde{u}^{res}_{\theta_\alpha}(o_\alpha))|o'_\alpha) \Big )
  \end{multline}
  \normalsize
  \State Update target networks: 
  \begin{equation}\nonumber 
      {\phi_{\alpha}^{\text{tar},i}} = \rho {\phi_{\alpha}^{\text{tar},i}} + (1-\rho) {\phi_{\alpha}^{i}}, \;\;\text{for}\;\; i = 1,2.
  \end{equation}
\EndFor
\If {federated update step}
    \State Compute federated average for critic and target \newline \hspace*{4.4em}$\text{for}\;\;\; i=1,2$.
%       \vspace{-0.05in}
%   \small
 \begin{gather*}\nonumber
        {\phi_{\text{fed}}^{i}} = \frac{1}{r}\sum_{\alpha=1}^{r} {\phi_{\alpha}^{i}}\;\;,\;\; {\phi_{\text{fed}}^{\text{tar},i}} = \frac{1}{r}\sum_{\alpha=1}^{r} {\phi_{\alpha}^{\text{tar},i}}
        % \;\;\;\text{for}\;\;\; i=1,2
    \end{gather*}
    % \State Compute federated average for target:
    %     \begin{equation}\nonumber
    %     {\phi_{\text{fed}}^{\text{tar},i}} = \frac{1}{r}\sum_{\alpha=1}^{r} {\phi_{\alpha}^{\text{tar},i}},\;\;\;\text{for}\;\;\; i=1,2
    % \end{equation}
\EndIf
\State Federated update:  ${\phi_{\alpha}^{i}} = {\phi_{\text{fed}}^{i}}$, and 
${\phi_{\alpha}^{\text{tar},i}} = {\phi_{\text{fed}}^{\text{tar},i}}$, \newline \hspace*{3.0em}for $i=1,2$, and for all $\alpha = 1,\cdots, r$.
\EndIf
\Until Convergence
\normalsize
\end{algorithmic} 
\label{alg_fedSAC}
\end{algorithm}
\normalsize
% \vspace{-.3 cm}
\setlength{\textfloatsep}{0.1cm}
\setlength{\floatsep}{0.1cm}

\section{Test Results}
\label{sec.results}
% \begin{figure}[t]
%   \centering
%     \includegraphics[width=\linewidth, height = 10 cm]{Figures/IEEE123_1.png}
%   \caption{\small{Modified IEEE 123-bus networked microgrid \cite{singhal2022consensus}}}
%   \label{f1}
%  \end{figure}
We perform the numerical experiments on a practical IEEE 123-bus test feeder system \cite{singhal2022consensus}.
% as shown in Figure \ref{f1}. 
The dynamic simulation is performed using GridLAB-D \cite{chassin2014gridlab} along with the HELICS \cite{palmintier2017design} based co-simulation interface as we intend to send action commands to the grid and utilize the observations as feedback. To perform RL training, we created a customized software interface using the OpenAI Gym platform that can communicate with the backend GridLAB-D/HELICS co-simulation engine. The network consists of three microgrids (MG) with the coupling via tie-lines. Each microgrid has 1 GFM and 2 GFL inverters with rating for GFM and GFL 600 kW and 350 kW, respectively. The total rating of the inverters is about 3900 kW, and the total peak load in the networked microgrid is about 3500 kW. All inverters have $1\%$ frequency droop and $5\%$ voltage droop values. The GFM inverters are connected at buses 1,4 and 7 for MGs 1, 2, and 3, while the GFL inverters at buses 2,3,4,5,8, and 9. Three-phase bus voltages of GFM and GFL inverters are considered as the observations for the underlying RL problem implying $|O| = 9\times 3 = 27$. Now considering multi-agent structure for the Fed-RL problem $|o_{\alpha}| = 3\times 3=9$, for $\alpha = 1,2,3$, as each MG has 3 inverters (1 GFM + 2 GFL).

\begin{figure}[t]
  \centering
    \includegraphics[width=\linewidth]{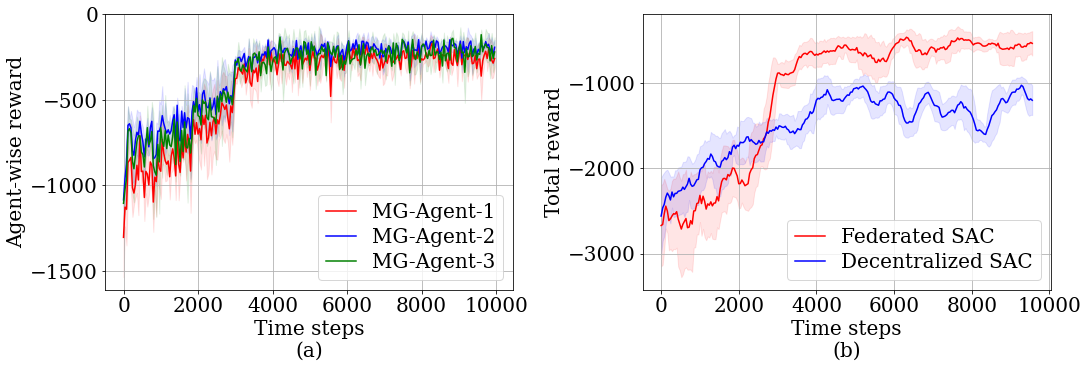}
  \caption{\small{(a) Agent-wise reward plot for Federated SAC training for 3 different seeds, (b) Comparison of Federated SAC and Multi-agent Decentralized SAC.}}
  \label{f2}
 \end{figure}
\begin{figure}[h]
  \centering
    \includegraphics[width=\linewidth]{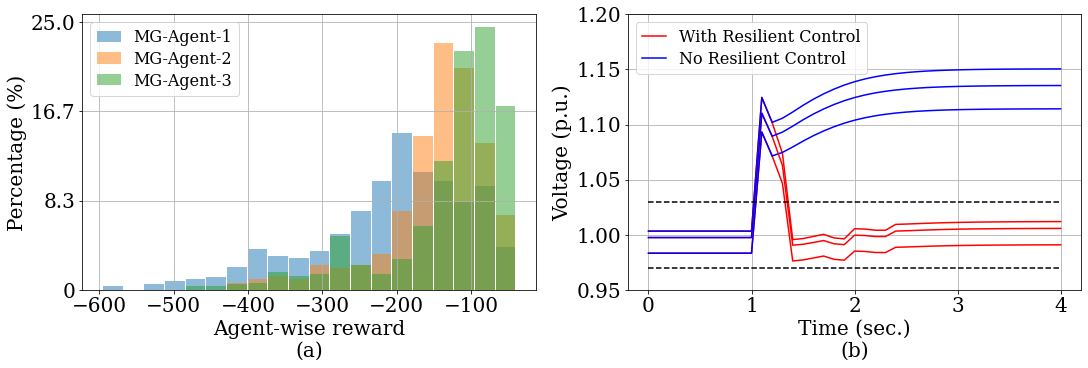}
  \caption{\small{(a) Histogram of Test Rewards, (b)  Voltage plot for Bus-1.}}
  \label{f3}
  \centering
    \includegraphics[width=\linewidth]{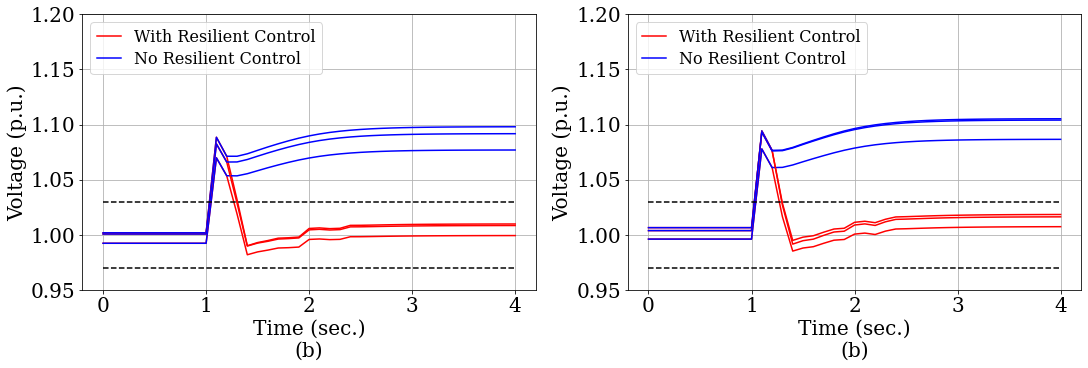}
  \caption{\small{(a) Voltage plot for Bus-4, (b) Voltage plot for Bus-7.
Dashed lines denote safety regions.}}
  \label{f4}
%   \vspace{-.4 cm}
 \end{figure}

We create a pool of adversarial perturbation at the voltage reference commands with a selected set of GFM inverters, i.e., for an individual run of the episode ($1$ episode $= 40$ time steps), one of the GFM inverter actuation has been made malicious. In the training phase, we created $7$ perturbation scenarios for the dynamic model. After that, the FedSAC algorithm is utilized as described in Alg. 2 and Alg. 3.{\color{black} Without loss of generality, as we attack only voltage reference commands, the microgrid-agent (MG-Agent) actions are selected as the voltage set point $V_i^{set}$ of GFM inverters of the respective MG. Both actor and critic structure of each MG-Agent have 2 hidden layers with 64 neurons per layer, and \textit{relu} activation function. The training parameters for SAC algorithm are chosen as: learning rate $=0.0003$, buffer size $= 1000000$, batch size $= 256$, $\rho=0.005$, $\gamma=0.99$. The federated learning is started after 100 time steps and is conducted at an interval of 10 time steps.} Fig. \ref{f2} (a) shows the training performance of FedSAC for three different microgrid agents with the mean and standard deviations plotted with 3 different seeds. Moreover, we experimented the proposed Fed-RL design against the fully decentralized architecture, and  Fig. \ref{f2} (b) shows a superior training performance for Fed-RL.
% where the agents are trained with the local observation-action in the dynamically coupled environment in a standard multi-agent training setup and 
% training performance as shown in Fig. \ref{f2} (b). 
To this end, we perform testing with $600$ different adversarial perturbation cases and collect the rewards for three MGs to plot the histogram as shown in Fig. \ref{f3} (a), where we can see the distribution has high probability mass for high reward values (perfect recovery), and low frequency with poor rewards near the tail, signifying high success rate. In the physical variables, Fig. \ref{f3}-\ref{f4} shows how the FedSAC has successfully recovered the voltages of the selected buses within the recovery margin, whereas the nominal microgrid model without the resilient controller fails, validating our design.

%   \begin{figure}[!htbp]
% \begin{minipage}[b]{.5\textwidth}
%   \centering
%   % include first image
%   \includegraphics[width=1.0\linewidth]{federated_reward.png}  
%   \caption{Reward plot for FedSAC training.}
%   \label{f2}
% %\vspace{4ex}
%   \end{minipage}
%   \begin{minipage}[b]{.5\textwidth}
%   \centering
%   % include second image
%   \includegraphics[width=1.0\linewidth]{fed_vs_decen.png}  
%   \caption{Reward comparison of Federated SAC and Decentralized SAC}
%   \label{f3}
%   \end{minipage}
%  \end{figure}

\vspace{-.1 cm}
\section{Conclusions}
% \vspace{-.1 cm}
We have proposed a novel vertical Fed-RL architecture with adversarial training for networked microgrids. The resilient control layer has been added in a hierarchical fashion on top of existing microgrid controls and tasked with recovering the microgrid dynamic voltage performance within conservative bounds. The federated soft actor-critic algorithm has shown to generate superior training performance than its decentralized multi-agent counterpart, and extensive testing with GridLAB-D/HELICS-OpenAI Gym platform shown to have a high probability toward successful recoveries. Our future research will investigate variations in the adversarial actions in the microgrids, with a focus on simulating some secondary level communication failures, and continue the development of novel resilient and secured learning algorithms.  

\bibliography{ref}
\bibliographystyle{IEEEtran}
%\bibliographystyle{abbrvnat}

% \section{Appendix}
% \subsection{Federated SAC Algorithm}
% We modified standard SAC algorithm from Stable Baselines \cite{raffin2021stable} to incorporate federated learning framework. The resulting algorithm is presented in Algorithm 3.

% \subsection{IEEE 123- Bus Network Description}
% A modified and fully inverter-based IEEE 123-node test feeder is used to demonstrate the impact of the proposed federated reinforcement learning-based controllers. The network consists of 9 utility-scale inverters, where 6 are GFL, and 3 are GFM inverters at nodes shown in Figure \ref{f1}. Each microgrid has 1 GFM and 2 GFL inverters with rating for GFM and GFL 600 kW and 350 kW, respectively. The total rating of the inverters is about 3900 kW, and the total peak load in the networked microgrid is about 3500 kW. All inverters have $1\%$ frequency droop and $5\%$ voltage droop values.
% \begin{figure}[htbp!]
%   \centering
%     \includegraphics[width=\linewidth]{Figures/IEEE123.png}
%   \caption{Modified IEEE 123-bus networked microgrid system.}
%   \label{f1}
%  \end{figure}

\end{document}